\documentclass[conference]{IEEEtran}

\IEEEoverridecommandlockouts
\PassOptionsToPackage{table,xcdraw}{xcolor}
\usepackage{cite}
\usepackage{amsmath,amssymb,amsfonts}
\usepackage[ruled,linesnumbered]{algorithm2e}
\usepackage{graphicx}
\usepackage{textcomp}
\usepackage{xcolor}
\usepackage{graphicx}
\usepackage{booktabs} 
\usepackage{multirow} 
\usepackage{url}
\def\Alg{QUARK }
\definecolor{myblue}{HTML}{BCCBE8}
\definecolor{mygreen}{HTML}{C6DFB8}
\definecolor{myorange}{HTML}{FCD5B4}
\begin{document}

\title{QUARK: Quantization-Enabled Circuit Sharing for Transformer Acceleration by Exploiting Common Patterns in Nonlinear Operations}

\author{
\IEEEauthorblockN{Zhixiong Zhao$^{3,\dagger}$, Haomin Li$^{1,\dagger}$, Fangxin Liu$^{1,2,*}$, Yuncheng Lu$^{3}$, Zongwu Wang$^{1,2}$, Tao Yang$^{4*}$ \\ Li Jiang$^{1,2}$, Haibing Guan$^{1}$}
\IEEEauthorblockA{1. School of Computer Science, Shanghai Jiao Tong University,\,\,\,2. Shanghai Qi Zhi Institute \\3. Nanyang Technological University, \,\,\,4. Huawei Technologies Co., Ltd}
\IEEEauthorblockA {*Corresponding Author \,\,\,\,\,\,\,\
zhixiong003@e.ntu.edu.sg, \,\, \{liufangxin, ljiang\_cs\}@sjtu.edu.cn
}
\thanks{This work was done when Zhixiong Zhao was an intern at Shanghai Jiao Tong University. $\dagger$ These authors contributed equally. This work is supported by the National Key Research and Development Program of China (2024YFE0204300), the National Natural Science Foundation of China (Grant No.62402311), and Natural Science Foundation of Shanghai (Grant No.24ZR1433700). Fangxin Liu and Tao Yang are the corresponding authors.}
}
\maketitle

\begin{abstract}
Transformer-based models have revolutionized computer vision (CV) and natural language processing (NLP) by achieving state-of-the-art performance across a range of benchmarks. However, nonlinear operations in models significantly contribute to inference latency, presenting unique challenges for efficient hardware acceleration. To this end, we propose QUARK, a quantization-enabled FPGA acceleration framework that leverages common patterns in nonlinear operations to enable efficient circuit sharing, thereby reducing hardware resource requirements. \Alg targets all nonlinear operations within Transformer-based models, achieving high-performance approximation through a novel circuit-sharing design tailored to accelerate these operations. Our evaluation demonstrates that \Alg significantly reduces the computational overhead of nonlinear operators in mainstream Transformer architectures, achieving up to a 1.96× end-to-end speedup over GPU implementations. Moreover, QUARK lowers the hardware overhead of nonlinear modules by more than 50\% compared to prior approaches, all while maintaining high model accuracy---and even substantially boosting accuracy under ultra-low-bit quantization. Code will be available at \url{https://github.com/Kishon-zzx/QUARK}.
\end{abstract}


\section{Introduction}
 In recent years, Transformer networks, have garnered significant attention in the field of Artificial Intelligence (AI), achieving remarkable results not only in tasks of Natural Language Processing (NLP) but also in Computer Vision (CV) through the Vision Transformer (ViT)\cite{dosovitskiy2021imageworth16x16words}. However, the traditional Transformer architecture heavily relies on the attention mechanism, which considerably increases both parameter count and computational complexity compared to earlier neural network models such as Convolutional Neural Networks (CNNs) or Recurrent Neural Networks (RNNs). (e.g.,ViT-L contains 307 million parameters and requires 190.7 billion floating-point operations (FLOP)\cite{AFP}).

\begin{figure}[t!]
    \setlength{\abovecaptionskip}{3pt}
    \setlength{\belowcaptionskip}{3pt}
    \vspace{-0.3cm}
    \centering
    \includegraphics[width=0.9\columnwidth]{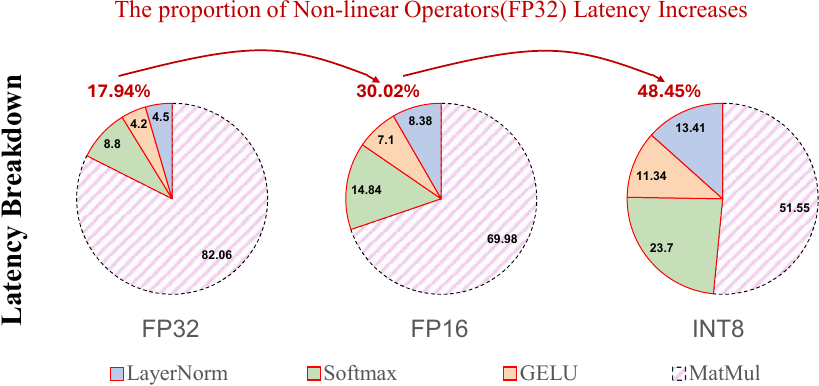}
    \vspace{-0.3cm}
    \caption{Latency Breakdown of ViT-Tiny on ZCU102.}
    \label{fig:latency}
    \vspace{-0.3cm}
\end{figure}

 Although architectural optimizations for models have been explored \cite{jouppi2017datacenter}, most transformer compression efforts focus on accelerating linear operators, which are conventionally regarded as the primary computational bottleneck in floating-point models\cite{kim2021bert,liu2024inspire}. In contrast, nonlinear operators have received limited attention despite their mathematical complexity and significant challenges in hardware acceleration, particularly as linear operators are quantized from 32-bit floating-point (FP32) to lower-bit integer arithmetic (e.g., 8-bit integer (INT8)). As illustrated in Figure \ref{fig:latency}, the latency contribution of nonlinear operators increases dramatically. These nonlinear operations are increasingly becoming critical bottlenecks in transformer optimization~\cite{liu2024spark}.

Recent studies on nonlinear operators primarily focus on optimizing individual or paired operators, overlooking the compounded computational impact of all nonlinear components (see Figure~\ref{fig:latency}). For instance, approaches like Softermax\cite{stevens2021softermax} and NN-LUT\cite{nnlut} target the Softmax operator exclusively, while FQ-ViT\cite{lin2023fqvitposttrainingquantizationfully} and SOLE\cite{wang2023sole} jointly optimize Softmax and LayerNorm. Although methods such as I-BERT\cite{kim2021bert} and I-ViT\cite{li2023vit} achieve full integer quantization for entire models, and they rely on quantization-aware training (QAT), requiring model retraining or fine-tuning to preserve post-quantization performance. Notably, as model scales increase, the associated retraining costs become prohibitively expensive. Consequently, striking an optimal balance between algorithmic efficiency and hardware constraints remains a critical challenge, particularly given the escalating size of transformer models.

We propose a novel hardware/software co-design framework \Alg to address two critical challenges: 1)Eliminating the reliance on floating-point operations in nonlinear operators through mathematical reformulation; 2)Reducing quantization errors caused by heterogeneity in activation distributions. Our key innovation lies in a shared sub-operator approximation framework, which enables joint optimization of common nonlinear functions such as Softmax, GELU, and LayerNorm. For Softmax, we propose an improved log-sum-exp algorithm\cite{hu2018efficient}, transforming computations from base-$e$ to base-2. This enables hardware implementation using shifters and adders, eliminating multiplication, division, and look-up tables (LUTs), and achieving an efficient 8-bit cache design. For GELU, we reformulate its computation as a combination of Softmax and ReLU\cite{glorot2011deep}, enabling hardware reuse of the Softmax structure alongside a simple ReLU circuit. For LayerNorm, we simplify its two data accesses to one, approximate the square root using an iterative method\cite{newton}, and transform the division operation logarithmically.
To address the challenges posed by diverse activation distributions, we propose a reordering-based group quantization scheme — a hardware-aware method that clusters channels during offline calibration based on distribution similarity. This approach adapts effectively to the unique characteristics of each layer, such as the power-law distribution in Softmax outputs, asymmetric activation in GELU, and inter-channel variation in LayerNorm. Group quantization scales are derived by shifting a base scale, enabling efficient cross-group alignment via parallel multipliers without requiring any floating-point conversions. In summary, our contributions are as follows:
\begin{itemize}
\item \textbf{Integer-Only Nonlinear Approximation:} We reformulate exponential, logarithmic, and division operations into low-cost shift-and-add arithmetic, eliminating the need for costly floating-point operations or large LUT-based approximations.
\item \textbf{Sub-Operator Sharing with Time-Division Multiplexing:} By exploiting common sub-operators (e.g., exponent, log) across Softmax, GELU, and LayerNorm, \Alg unifies them into a single reusable hardware block, significantly reducing resource usage and power consumption.
\item \textbf{Reorder-Based Group Quantization:} We propose a novel group quantization mechanism that leverages offline reordering and scaled integer alignment to effectively handle diverse output distributions. This approach prevents significant degradation in accuracy under ultra-low-bit settings and simplifies hardware design for multi-group data alignment.
\end{itemize}

\section{Related Work}
\paragraph{Optimization of Nonlinear Operators}
Prior works primarily target Softmax optimization. Approaches like A3\cite{A3} and NN-LUT\cite{nnlut} leverage LUTs or piecewise linear fitting to approximate exponentials but rely on costly division units. SpAtten\cite{wang2021spatten} and ELSA\cite{ham2021elsa} perform quantized Softmax via floating-point units, which are underutilized and inefficient. CORDIC-based designs\cite{spagnolo2021aggressive} improve hardware efficiency but sacrifice accuracy.
I-BERT\cite{kim2021bert} introduces a second-order integer-only GELU approximation, reducing arithmetic complexity compared to floating-point-based error functions and log-sum-exp methods\cite{10595882}. However, it still requires 32-bit intermediate caching, incurring high memory overhead. SpAtten also trims input sizes in LayerNorm for performance gains. I-ViT\cite{li2023vit} adopts shift-adder–based integer approximations but still depends on expensive divisions and retraining to preserve accuracy.

\paragraph{Model Quantization}
Quantization has emerged as a crucial technique for reducing model size and computational requirements in LLMs. Current approaches primarily fall into two categories: QAT and PTQ. QAT adapts models to quantization noise through retraining, while PTQ directly converts FP32 models to low-bit formats without requiring training data, offering greater practicality for Transformers.
However, ultra-low-bit ($\leq$4-bit) quantization remains challenging due to nonlinear operator heterogeneity. Existing methods like Qbert\cite{shen2020q} and VS-Quant\cite{dai2021vs} apply uniform channel grouping, neglecting inter-channel dynamic range variations and introducing large errors. PEG\cite{bondarenko2021understanding} addresses global statistical divergence but ignores local nonlinear behavior. Quantformer\cite{wang2022quantformer} explores search-based grouping in QAT but suffers from a high overhead design, limiting deployment efficiency.

\begin{figure}[t!]
    \setlength{\abovecaptionskip}{3pt}
    \setlength{\belowcaptionskip}{3pt}
    \vspace{-0.3cm}
    \centering
    \includegraphics[width=\columnwidth]{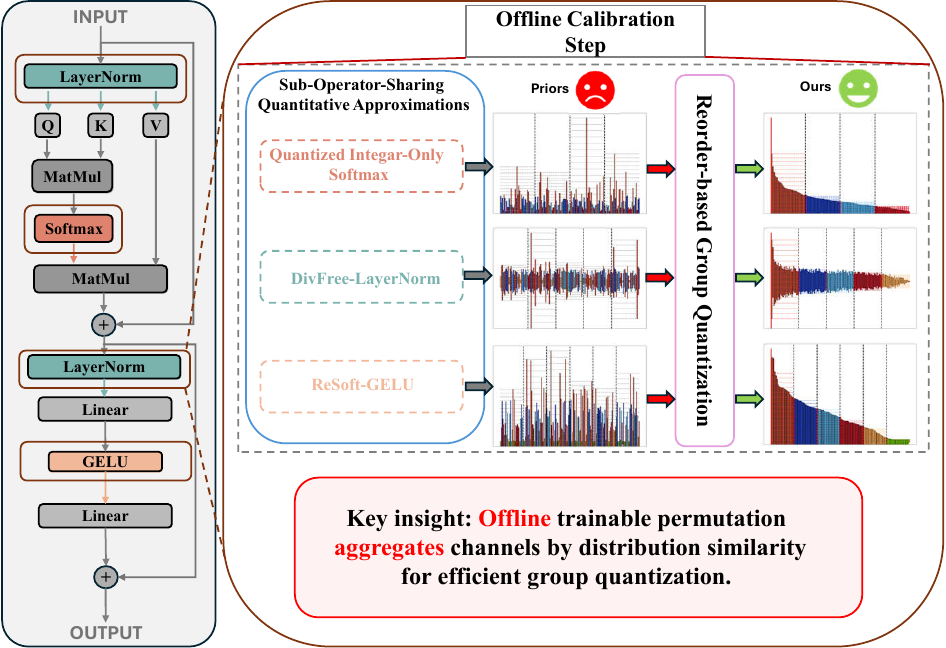}
    \caption{Overview of \Alg Software.}
    \label{fig:overview}
\end{figure}

\section{\Alg Algorithm}
In QUARK, we introduce a novel sub-operator-sharing approximation framework specifically designed for three key nonlinear operators in Transformers, enabling efficient deployment of quantized models (Figure \ref{fig:overview}). 
First, we detail our sub-operator-sharing approximation methodology. Subsequently, we analyze the resulting activation distributions and derive our reordering-based group quantization approach.
\subsection{Sub-Operator-Sharing Quantitative Approximations}
\textbf{Quantized Integer-Only Softmax.} The Softmax operator is widely used as a normalization function in Transformers and can be defined as follows:
\begin{equation}
     \text{Softmax}(x_i) = \frac{\exp(x_i)}{\sum_{j=1}^n \exp(x_j)}
\end{equation}
Due to the nonlinear nature of Softmax, it cannot be directly quantized like linear operators, and integer-only logic typically does not support exponential arithmetic. Additionally, overflow may occur. To this end, we restrict the range of the exponential operation and eliminate division by using approximate logarithmic and exponential functions, defined as follows:
\begin{equation}
\begin{gathered}  
    \text{Softmax}(x_i) = \exp\{x_i - \ln[\sum_{j=1}^n \exp(x_j)]\} \\
    = \exp\{(x_i- x_\text{max})- \ln[\sum_{j=1}^n \exp(x_j- x_\text{max})]\}
\end{gathered}
\end{equation}
where $x_{\text{max}} = \max\{ x_1, x_2, \dots, x_n \}$ and ${x_i}^{'} = x_i-x_\text{max}$. The term $x_i^{'}$ simplifies the expression and is always non-positive. For the exponential operation, we express $\log_2 e$ as a binary approximation, $\text{(1.0111)}_b$, to convert from base $e$ to base 2, using only shifters and adders. This reduces hardware complexity. Since the exponential term of 2 may contain a fractional part, we separate it into integer and fractional components:
\begin{equation}
\begin{gathered}
    \exp(x_i^{'}) = 2^{({x_i}^{'} \cdot \log_2e)} \\
    \approx 2^{[{x_i}^{'}+({x_i}^{'} \gg 1) - ({x_i}^{'} \gg 4) ]} = 2^{q_{I} \cdot q_{F} }
\end{gathered}
\end{equation}
where $q_{I}=\lceil {x_i}^{'}+({x_i}^{'} \gg 1) - ({x_i}^{'} \gg 4) \rceil$, represents the integer part, which corresponds to the non-positive exponent of 2 rounded upwards, and $q_{F} = {x_i}^{'}+({x_i}^{'} \gg 1) - ({x_i}^{'} \gg 4)-q_{I} \in (-1,0] $, represent the fractional part. For low-cost computation, we shift the integer part directly, while the fractional part, within the range of (-1,0], is approximated using a second-order polynomial as follows:
\begin{equation}
    2^{q_{I} \cdot q_{F}} = 2^{q_{F}} \ll q_{I} 
\end{equation}
\begin{equation}
    2^{q_{F}} \approx 0.1713q_{F}^2+0.6674q_{F}+0.998
\end{equation}
In the logarithm computation, we convert the natural logarithm to a base-2 logarithm for more efficient computation. Since $\ln 2$ is approximated as $\text{(0.1011)}_b$, we can use shifts and additions to convert the logarithm. The logarithm of $S{x_i} \cdot I_{x_i}^{' }$ is then approximated as:
\begin{equation}
\begin{gathered}
    \ln({x_i}^{'}) = ln2 \cdot \log_2({x_i}^{'}) \\
    \approx  \log_2({x_i}^{'}) - [\log_2({x_i}^{'}) \gg 2] -[\log_2({x_i}^{'}) \gg 4]
\end{gathered}
\end{equation}
The term $\log_2({x_i}^{'})$ can be reformulated by expressing${x_i}^{'}$ as the product of an integer power of 2 and a positive value within the range [1, 2). Using this decomposition and the mathematical properties of the logarithm function, the original logarithmic expression can be separated into the sum of two terms as follows:
\begin{equation}
    {x_i}^{'} = 2^{q_{M}}\cdot q_{N}, \; \text{where} \, q_{N} \in [1, 2)
\end{equation}
Then, we can rewrite the logarithmic expression as:
\begin{equation}
\begin{gathered}
    \log_2({x_i}^{'}) = \log_2(2^{q_{M}}\cdot q_{N}) = q_{M} + \log_2(q_N)
\end{gathered}
\end{equation}
Here, $q_{M}$ represents the position of the most significant bit (MSB) of ${x_i}^{'}$ in binary form, and $q_{N}$ is the normalized factor. Since $q_{N}$ lies in the range [1, 2), we approximate $\log_2(q_{N})$ using a second-order polynomial as follows::
\begin{equation}
    \log_2(q_{N}) \approx -0.3369q_{N}^2+1.995q_{N}-1.65
\end{equation}
Algorithm \ref{alg:Softmax} summarizes the integer-only Softmax flow.

\begin{algorithm}[t]
    \setlength{\abovecaptionskip}{3pt}
    \setlength{\belowcaptionskip}{3pt}
    \caption{Quantized Integer-Only Softmax.}
    \label{alg:Softmax}
    
    \let\oldnl\nl
    \newcommand{\nonl}{\renewcommand{\nl}{\let\nl\oldnl}}
    
    \textbf{Input:} Integer input $x_{\text{in}}$ \\
    \textbf{Output:} Integer output $x_{\text{out}}$ 
    \BlankLine

   \SetKwProg{Fn}{Function}{:\hfill  $\triangleright$  $x_{\text{in}}^{'}=x_{\text{in}}-x_{\text{in}_\text{max}}$}{end}
    
    \Fn{\textnormal{\textbf{Appro-Exp}(}$x_{\text{in}}^{'}$\textnormal{\textbf{)}}} 
    {
        
        
        $x_{shift} \leftarrow x + (x \gg 1) - (x \gg 4)$ \hfill $\triangleright$ $x \cdot \log_2 e$
        $q_{I} \leftarrow  \lceil x_{shift} \rceil$ \hfill $\triangleright$ $\text{Integer part}$ \\
        $q_{F} \leftarrow x_{shift} - q_{I}$ \hfill $\triangleright \text{fractional part}$
        $x_a \leftarrow 0.1713q_{F}^2+0.6674q_{F}+0.998$ \hfill $\triangleright \text{Eq. 5}$\\
        $x_{exp} \leftarrow x_a \ll q_{I}$ \hfill $\triangleright \text{Eq. 4}$\\
        \Return $x_{exp}$ \hfill $\triangleright x_{exp} \approx \exp(x_{\text{in}}^{'}) $
        }
    \BlankLine
    
    \SetKwProg{Fn}{Function}{:\hfill  $\triangleright$ $x_{\text{in}}^{'}=x_{\text{in}}-x_{\text{in}_\text{max}}$}{end}
    
    \Fn{\textnormal{\textbf{Appro-Ln}(}$x_{\text{in}}^{'}$\textnormal{\textbf{)}}} 
    {
        
        
        $q_M \leftarrow MSB(log_2(x_{\text{in}}^{'})$ \\
        $q_N \leftarrow  x_{\text{in}}^{'} \gg q_a$ \hfill $\triangleright$ $\text{Eq. 7}$\\
        $x_a \leftarrow -0.3369q_{N}^2+1.995q_{N}-1.65$ \hfill $\triangleright$ $\text{Eq. 9}$\\
        $x_{ln} \leftarrow  x_a - (x_a \gg 2) - (x_a \gg 4)$ \hfill $\triangleright$ $I \cdot \ln2$\\
        \Return $x_{ln}$ \hfill $\triangleright x_{ln} \approx \ln(x_{\text{in}}^{'})$
        }

    \BlankLine
    
    \SetKwProg{Fn}{Function}{:}{end}
    
    \Fn{\textnormal{\textbf{IntDiv-Free Softmax}(}$x_{\text{in}}$\textnormal{\textbf{)}}} 
    {
        
        
        $x_{\text{in}}^{'} \leftarrow x_{in}-max(x_{in})$ \\
        $x_{exp} \leftarrow  \textbf{Appro-Exp}(x_{\text{in}}^{'})$ \\
        $x_{ln} \leftarrow  \textbf{Appro-Ln}(\sum x_{\text{in}}^{'})$ \\
        $x_0 \leftarrow x_{exp} - x_{ln}$\\
        $x_{\text{out}} \leftarrow  \textbf{Appro-Exp}(x_0)$ \hfill $\triangleright$ $\text{Eq. 2}$\\
        
        \Return $x_{\text{out}}$ \\
        \hfill $\triangleright x_{\text{out}} \approx \text{Softmax}(x_{\text{in}})$
        }     
        
\end{algorithm}

\textbf{Quantized Integer-Only GELU with ReLU and Softmax.}
GELU~\cite{gelu} is a nonlinear activation function used in Transformer models, defined as:
\begin{equation}
\begin{gathered}
    \text{GELU}(x_i) = x_i \cdot \frac{1}{2} \left[ 1 + \text{erf}\left( \frac{x_i}{\sqrt{2}} \right) \right],\\
    \text{where} \; \text{erf}(x_i) = \frac{2}{\sqrt{\pi}} \int_0^{x_i} \exp(-t^2) \, dt
\end{gathered}
\end{equation}
Here, $erf$ is the error function, which is computationally inefficient. Based on the findings in \cite{gelu}, the GELU operator can be approximated using a Sigmoid function as follows:
\begin{equation}
\text{GELU}(x_i)  \approx x_i \cdot \sigma(1.702x_i) 
\end{equation}
where $\sigma(\cdot)$ is the Sigmoid function. Since the Sigmoid function requires floating-point arithmetic, it is not suitable for integer-only quantization. However, we observe that the Sigmoid function has a form identical to the first output element of the Softmax function when the input is a binary vector. Thus, we can approximate the Sigmoid using a binary Softmax: 

\begin{equation}
\begin{gathered}
    \sigma(1.702x_i) =  \frac{1}{1+e^{-1.702x_i}}\\
    = \frac{e^0}{e^0+e^{-1.702x_i}}
    =\text{Softmax}_2^1([0,-1.702x_i])
\end{gathered}
\end{equation}
leading to the approximation:
\begin{equation}
    \text{GELU}(x_i) \approx 1.702x_i \cdot \text{Softmax}_2^1([0,-1.702x_i])
\end{equation}

\begin{figure}[t!]
    \setlength{\abovecaptionskip}{3pt}
    \setlength{\belowcaptionskip}{3pt}
    \vspace{-0.5cm}    
    \centering
    \includegraphics[width=0.9\columnwidth]{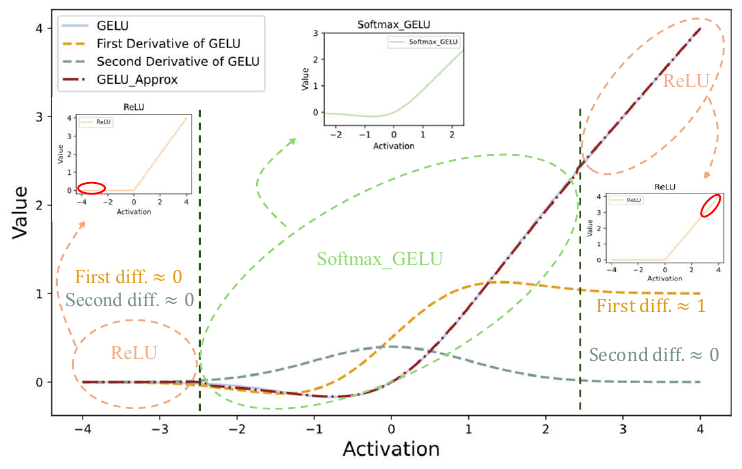}
\vspace{-0.3cm}
    \caption{Characteristics of GELU Operator in Transformers.}
    \label{fig:GELU}
    \vspace{-0.6cm}
\end{figure}

Here,$(\cdot)_2^1$ represents the first output element with a binary vector as input. However, we observe that this approximation introduces a significant amount of redundant computation when applied to all activation values, as it fails to consider the specific characteristics of the activation values. As shown in figure~\ref{fig:GELU}, the second-order derivative of the GELU operator approaches zero for activation values with large absolute magnitudes, making it well-approximated by a linear function. Additionally, the first-order derivative in these regions closely resembles that of the ReLU operator. Based on these observations, we employ the ReLU operator for larger activation values to enhance computational efficiency while preserving accuracy. For the remaining intervals, we retain the approximation algorithm described in $\text{Eq. 13}$. Since this algorithm is applied only to specific intervals, it does not significantly contribute to the overall computational overhead, which ReLU operator as follows:

\begin{equation} 
    \text{GELU}(x_i) \approx \text{ReLU}(x_i) = \max(0,x_i) 
\end{equation}

For selecting the approximation intervals, the boundary points of the ReLU approximation interval are determined by the activation values where the absolute errors of $\text{Eq. 13}$ and $\text{Eq. 14}$, when compared to $\text{Eq. 10}$ are equal. The ReLU approximation interval is consequently defined as $ (-\infty, -2.4] \cup [2.4, \infty) $, ensuring computational efficiency while maintaining accuracy in these regions. 

    
    

    
        
        
    
    
         
        
            
       
        

\textbf{Integer-only LayerNorm without Division}
LayerNorm is a widely used technique in Transformer models to enhance training stability and model performance\cite{layernorm}. The primary objective of the LayerNorm operator is to normalize the output of each layer to have zero mean and unit variance, thereby mitigating the distributional shifts in the inputs to each layer during training. In the context of Transformers, LayerNorm normalizes the inputs to each layer along the hidden feature dimension to achieve a unit variance. The LayerNorm operation in Transformer models is formally expressed as follows:
\begin{equation}
    \text{LayerNorm}(x_i) = \frac{x_i-\text{Mean}(x_i)}{\sqrt{\text{Var}(x_i)}}
\end{equation}
where $\text{Mean}(\cdot)$ and $\text{Var}(\cdot)$ represent the mean and variance of the input across the feature dimension.

This process involves several nonlinear operations, including division, squaring, and square roots. Given the rapid rate of change of $\text{Mean}(\cdot)$ and $\text{Var}(\cdot)$ during inference, these values must be computed dynamically at runtime. To minimize data access overhead, we first simplify the variance calculation mathematically, reducing the need for two full data accesses to just one, as follows:
\begin{equation}
\text{Var}(x_i) = \mathbb{E}\{[x_i - \mathbb{E}(x_i)]^2\} = \mathbb{E}(x_i^2) - [\mathbb{E}(x_i)]^2
\end{equation}
where $\mathbb{E}(\cdot)$ is the mean value. Previous work\cite{kim2021bert} optimized the square root operator using an iterative approach; however, it did not address the large number of division operations involved, both in the iterative process and within the LayerNorm operator itself. To overcome this, we optimize the LayerNorm operator by employing shifting and logarithmic division which is the same as Softmax as follows :
\begin{equation}
     x_{i+1} = \left(x_i + \left\lfloor \frac{\text{Var}(x)}{x_i} \right\rfloor \right) / 2 
    = \left(x_i + \left\lfloor \frac{\text{Var}(x)}{x_i} \right\rfloor \right) \gg 1
\end{equation}
\begin{equation}
    \frac{\text{Var}(x)}{x_i} = \exp(\ln{[ \frac{\text{Var}(x)}{x_i} ] }) = \exp(\ln[\text{Var}(x)] - \ln(x_i))
\end{equation}
where $x_i$ is the result of the \( i \)-th iteration, initialized as $x_0 = 2^{\lfloor \text{bit}(\text{Var}(x)) / 2 \rfloor}$. This iterative method converges within ten iterations for any INT8 input, avoiding nonlinear operations like division. 

\subsection{Activation Distribution after Quantized Nonlinear Approximations}
\begin{figure}[t!]
    \setlength{\abovecaptionskip}{3pt}
    \setlength{\belowcaptionskip}{3pt}
    \vspace{-0.3cm}
    \centering
    \includegraphics[width=\columnwidth]{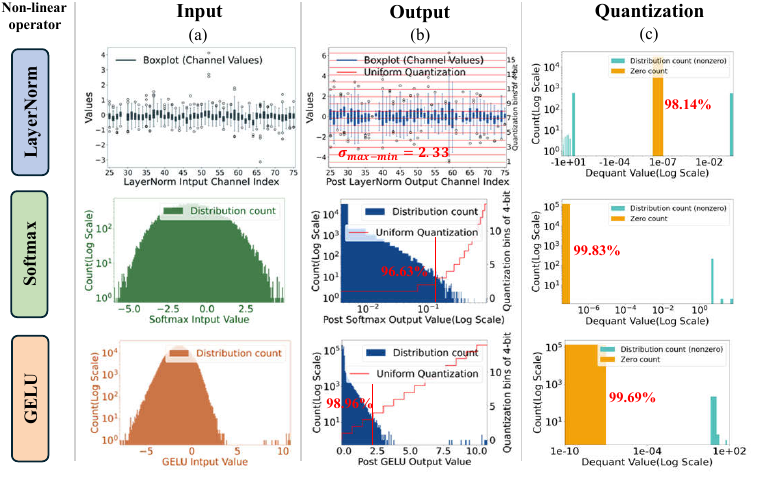}
    \vspace{-0.5cm}
    \caption{Visualization of post-nonlinear activations from the 6th layer in DeiT-T, illustrating key Transformer quantization challenges: LayerNorm's inter-channel variance, Softmax's zero-collapsing heavy-tailed distribution, and GELU's asymmetric activation range that challenges conventional symmetric quantization schemes.}
    \label{fig:distribution}
    \vspace{-0.3cm}
\end{figure}

In the previous section, we performed quantization approximation on three nonlinear operators in Transformers. However, as shown in Figure \ref{fig:distribution}, the activation distributions of these approximated nonlinear operators exhibit significant non-uniformity. Directly applying identical quantization parameters across entire layers may introduce substantial quantization errors. Through detailed analysis of post-nonlinear activation distributions: 1) \textbf{LayerNorm} outputs show severe inter-channel variance - some channels contain prominent outliers where maximum/minimum values exceed those of other channels by orders of magnitude. While per-channel quantization could mitigate this issue, it introduces prohibitive computational overhead. 2) \textbf{Softmax} transforms attention scores into (0,1)-bounded probabilities, yet its outputs follow a power-law distribution. In DeiT-S, 96.63\% of values reside below 0.1, while the remaining 3.37\% encode crucial token dependencies. 3) \textbf{Post-GELU} activations demonstrate asymmetric characteristics: negative values cluster near zero forming a sharp peak ($\mathbb{E}[x] \approx -0.1$, $\text{Var}(x) < 0.01$), whereas positive values follow a heavy-tailed distribution extending towards higher magnitudes ($\mathbb{E}[x] \geq 0.8$, $\text{Var}(x) \propto x^2$).

\subsection{Reorder-based Group Quantization}
We propose QUARK, a novel quantization method that employs a reordered grouping quantization scheme specifically designed for post-nonlinear operator activations. This approach adapts to the distinctive characteristics of different nonlinear operators through a unified quantization framework, which performs channel-wise statistical analysis to guide the sorting and grouping of activations. Notably, to eliminate the computational overhead of online reordering, we strategically integrate the reordering process into the model during the offline calibration phase, thereby enabling seamless deployment without runtime penalties.

\begin{figure*}[t!]
    \setlength{\abovecaptionskip}{3pt}
    \setlength{\belowcaptionskip}{3pt}
    \vspace{-0.8cm}
    \centering
    \includegraphics[width=0.9\textwidth]{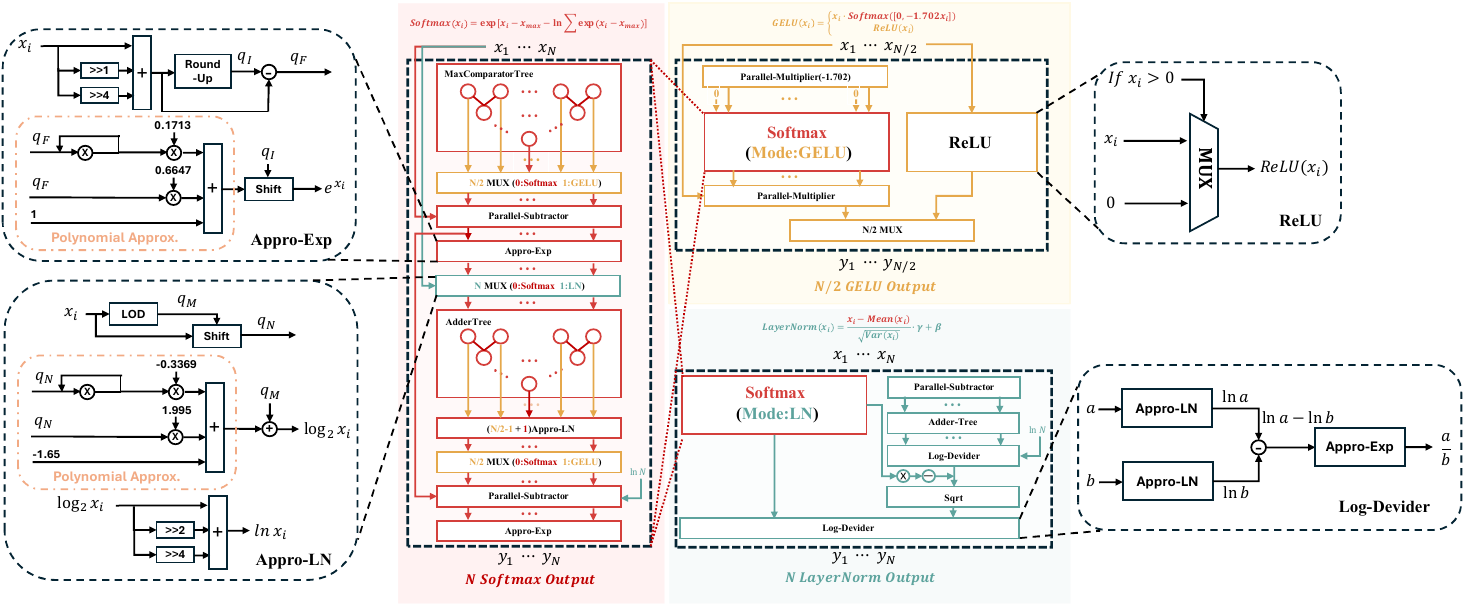}
    \caption{The shared hardware components and reuse pathways among Softmax, GELU, and LayerNorm.}
    \label{fig:nonlinear}
    \vspace{-0.5cm}
\end{figure*}

\textbf{Group Size Allocation}
We observe that activations after nonlinear operators in different layers exhibit significant channel-wise distribution divergence, indicating heterogeneous sensitivity to quantization noise across layers. This finding suggests that using a fixed grouping strategy for all layers may lead to suboptimal quantization performance. To address this, we propose optimizing layer-wise group sizes under a bitwise operations (BOP) constraint to minimize the distribution discrepancy between the quantized and full-precision models. Formally, given a target BOP budget $N_{\text{bop}}$, we formulate the group allocation problem as an integer linear programming (ILP) optimization:

\begin{equation}
\begin{aligned}
\min_{\{g_l\}} \sum_{l=1}^{L} D_{\text{KL}}\left(P_{\text{FP}}(a_l) \,\|\, P_{\text{Q}}(a_l; g_l)\right)
\quad \text{s.t.}  \sum_{l=1}^{L} B_l(g_l) \leq N_{\text{bop}}, \\
\quad g_l \in \mathbb{Z}^+, \, l=1,2,\dots,L
\end{aligned}
\end{equation}
where $L$ denotes the total number of layers in the network, $g_l \in \mathbb{Z}^+$ is the number of groups used for quantizing activations in layer $l$, $P_{\text{FP}}(a_l)$ and $P_{\text{Q}}(a_l; g_l)$ represent the full-precision and quantized activation distributions of layer $l$, respectively, and $B_l(g_l) = C_l \cdot b \cdot \log_2 g_l$ models the BOP cost for layer $l$ when using $g_l$ groups, with $C_l$ being the number of channels in layer $l$ and $b$ the quantization bit-width. The constraint ensures that the total BOP across all layers does not exceed the target budget $N_{\text{bop}}$.

\textbf{Offline Reordering}
To enable group quantization by clustering channels with similar statistical characteristics, we first need to reorder activation channels based on their statistical properties. However, performing such reordering during inference introduces additional latency, especially in large-scale models with a large number of channels. Moreover, storing both the original and reordered activation tensors leads to increased memory overhead. To minimize the cost of reordering, we introduce a trainable permutation matrix $R$, constrained to be orthogonal, and fuse it into adjacent operations (e.g., linear layers or LayerNorm) during the offline calibration stage. To ensure numerical equivalence, we invert the permutation matrix and absorb it into the weight matrix of the current layer. The transformation is expressed as:
\begin{equation} 
Y = (X \cdot R) \cdot (R^{-1} \cdot W) = X \cdot W 
\end{equation}
where $R$ is a diagonal matrix and $R^{-1}$ denotes its element-wise inverse along the diagonal.
Since the input $X$ typically originates from a preceding linear operation (such as a linear layer or LayerNorm), the permutation matrix $R$ can be fused offline into the parameters of the preceding layer. This design ensures that no significant computational overhead is introduced during inference. As an example, consider reordering the activations after the Softmax operation in the attention mechanism. In this case, the reordering matrix $R_{vo}$ can be fused into the adjacent projection matrix $W_V$, while its inverse transpose can be absorbed into the subsequent output projection matrix $W_O$. The output of the attention block becomes:
\begin{equation} 
Y_O = \left[ \left( \mathrm{Attn} \cdot V \right) W_V R_{vo} + b_V R_{vo} \right] R_{vo}^\top W_O + b_O
\end{equation}

\textbf{Intra-group Quantization and Cross-group Alignment} 
We divide the activations into groups $G_1, G_2, \dots, G_n$ according to Equation (19). Within each group, normal elements are quantized using a uniform scalar quantizer, while outliers are clamped or frozen to preserve critical information. For a normal element $x \in G_1$, the quantized value is computed as: 
\begin{equation} 
x_q = \left\lfloor \frac{x}{\Delta_{R_1}} + \frac{1}{2} \right\rfloor \end{equation}
For outliers $y \in G_1$ that exceed the threshold, the quantized value is mapped directly to the quantization boundary: 
\begin{equation} 
y_q = \mathrm{sign}(y) \times Q_{\max} 
\end{equation}
The complete quantization rule can be described as: 
\begin{equation}
z_q = 
\begin{cases} 
\left\lfloor \frac{z}{\Delta_{\mathcal{R}_1}} + \frac{1}{2} \right\rfloor & \text{if } |z| \leq \text{threshold} \\
\text{sign}(z) \times Q_{\max} & \text{if } |z| > \text{threshold}
\end{cases}
\end{equation}
This approach preserves important outlier information while enabling efficient quantization of the remaining values. To support hardware-friendly processing across different groups, we assign a group-wise quantization step: 
\begin{equation} 
\Delta_{R_i} = 2^{k_i} \cdot \Delta_{R_1}, \quad i = 1, 2, \dots, n 
\end{equation} 
where $k_i$ is the scaling factor for group $G_i$, and $\Delta_{R_1}$ is the reference step size for group $G_1$.
This formulation simplifies alignment during matrix multiplications. For example, aligning quantized values $a_q \in G_1$ and $b_q \in G_2$ results in: 
\begin{equation} 
a_q \cdot \Delta_{R_1} + b_q \cdot \Delta_{R_2} = a_q + b_q \cdot 2^{k_2 - k_1} \cdot \Delta_{R_1} 
\end{equation}
Extending this to $n$ groups, the generalized alignment is: 
\begin{equation} 
a_q \cdot \Delta_{R_1} + \sum_{i=2}^{n} b_{q_i} \cdot \Delta_{R_i} = (a_q + \sum_{i=2}^{n} b_{q_i} \cdot 2^{k_i - k_1}) \cdot \Delta_{R_1} 
\end{equation}
Here, $b_{q_i}$ is a quantized value from group $G_i$, and $k_i$ is its corresponding scaling factor. This strategy ensures efficient cross-group alignment with minimal computational overhead.

\section{\Alg Hardware}
\Alg is a pluggable nonlinear unit that connects to the PE array via a shared buffer, making it compatible with existing accelerators and programmable platforms (e.g., FPGAs). This design enables fast Transformer inference while preserving hardware flexibility.

\subsection{Sub-Operator Sharing Unit}

\Alg improves efficiency and reduces hardware cost by enabling sub-operator reuse across nonlinear functions such as Softmax, GELU, and LayerNorm. A key innovation is the reformulation of GELU into a binary Softmax-like structure, allowing both to share an exponential computation module, as illustrated in Figure~\ref{fig:nonlinear}.
To minimize computational overhead, complex functions (e.g., $\exp$, $\log$, $\div$) are approximated using lightweight adder- and shifter-based logic. This enables a unified arithmetic backend based on adders and shift registers to support multiple nonlinear operators.

Given that these functions are invoked at different stages of the Transformer pipeline---e.g., attention, activation, normalization---\Alg applies time-division multiplexing (TDM). TDM reuses hardware units sequentially across pipeline stages, minimizing area and energy overhead without performance compromise.
Figure~\ref{fig:nonlinear} summarizes the shared arithmetic paths across operators enabled by this design.

\subsection{Group Quantization Unit}
\begin{figure}[t!]
    \setlength{\abovecaptionskip}{3pt}
    \setlength{\belowcaptionskip}{3pt}
    \vspace{-0.2cm}
    \centering
    \includegraphics[width=\columnwidth]{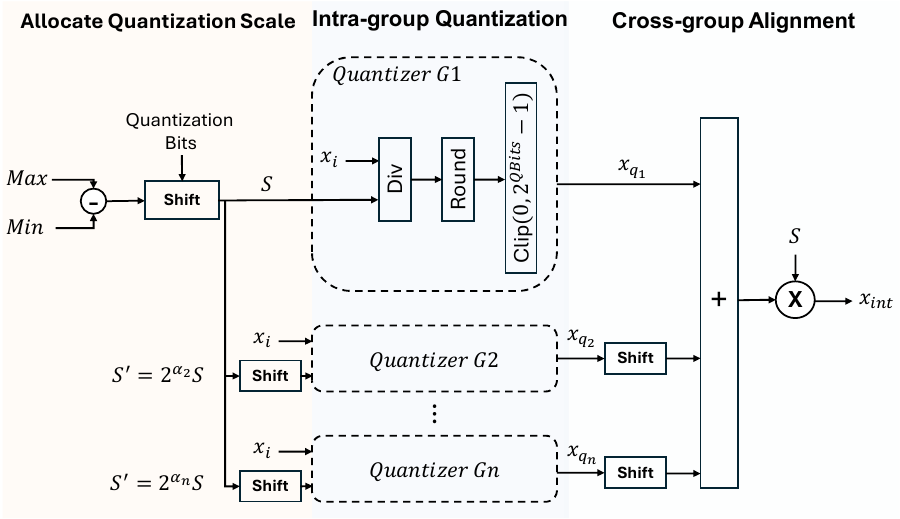}
    \vspace{-0.5cm}
    \caption{Architecture of Group Quantization Unit.}
    \label{fig:group}
    \vspace{-0.4cm}
\end{figure}

\begin{table*}[h!]
    \setlength{\abovecaptionskip}{3pt}
    \setlength{\belowcaptionskip}{3pt}
    \vspace{-0.6cm}
\footnotesize
\centering
\caption{Comparison with state-of-the-art (SOTA) methods on ImageNet. Our method achieves minimal accuracy drop while meeting both constraints of no retraining and integer-only inference. Notably, under ultra-low bit-width settings (4/4), it significantly outperforms existing methods in accuracy. (``Prec. (W/A)'' indicates the bit-widths for weights and activations, respectively.)}
\vspace{-0.2cm}
\begin{tabular}{cccccccccccc}
\hline
\textbf{Method} & \textbf{Prec.(W/A)} & \textbf{Training-free} & \textbf{Int.-only} & \textbf{ViT-S} & \textbf{ViT-B}
& \textbf{DeiT-T} & \textbf{DeiT-S} & \textbf{DeiT-B} & \textbf{Swin-T} & \textbf{Swin-S}
& \textbf{Swin-B} \\ \hline
Full-Precision &32/32 &- &- &81.39 &84.54 &72.21 &79.85 &81.80 &81.35 &83.23 &85.27\\ \hline
I-BERT\cite{kim2021bert} &8/8 &$\times$ &$\checkmark$ &80.47 &83.70 &71.33 &79.11 &80.79 &80.15 &81.86 &- \\
I-ViT\cite{li2023vit} &8/8 &$\times$ &$\checkmark$ &81.27 &84.76 &72.24 &80.12 &81.74 &81.50 &83.01 &- \\
FQ-ViT\cite{lin2021fq} &8/8 &$\checkmark$ &$\times$ &78.68 &82.76 &70.92 &78.44 &81.12 &79.97 &82.38 &82.53\\
PTQ4ViT\cite{yuan2022ptq4vit} &8/8 &$\checkmark$ &$\times$ &81.00 &84.25 &71.72 &79.47 &81.48 &81.23 &83.10 &85.14\\
RepQ-ViT\cite{li2023repq} &8/8 &$\checkmark$ &$\times$ &81.19 &84.39 &72.03 &79.68 &81.77 &81.19 &83.05 &85.12\\
 \rowcolor{green!15} 
\textbf{QUARK(ours)} &8/8 &$\checkmark$ &$\checkmark$ &80.72 &83.78 &71.29 &79.40 &81.52 &81.06 &82.81 &85.03 \\ \hline
FQ-ViT\cite{lin2021fq} &6/6 &$\checkmark$ &$\times$ &4.26 &0.10 &58.66 &45.51 &64.63 &48.63 &66.50 &52.09\\
PTQ4ViT\cite{yuan2022ptq4vit} &6/6 &$\checkmark$ &$\times$ &78.63 &81.65 &69.68 &76.28 &80.25 &80.47 &82.38 &84.01\\
RepQ-ViT\cite{li2023repq} &6/6 &$\checkmark$ &$\times$ &80.43 &83.62 &70.76 &78.90 &81.27 &80.89 &82.79 &84.57\\
\rowcolor{green!15} 
\textbf{QUARK(ours)} &6/6 &$\checkmark$ &$\checkmark$ &80.12 &83.28 &70.29 &78.60 &81.42 &81.08 &82.93 &84.78 \\ \hline
FQ-ViT\cite{lin2021fq} &4/4 &$\checkmark$ &$\times$ &0.10 &0.10 &0.10 &0.10 &0.10 &0.10 &0.10 &0.10\\
PTQ4ViT\cite{yuan2022ptq4vit} &4/4 &$\checkmark$ &$\times$ &42.57 &30.69 &36.96 &34.08 &64.39 &73.48 &76.09 &74.02\\
RepQ-ViT\cite{li2023repq} &4/4 &$\checkmark$ &$\times$ &65.05 &68.48 &57.43 &69.03 &75.61 &72.39 &79.45 &78.32\\
 \rowcolor{green!15} 
\textbf{QUARK(ours)} &4/4 &$\checkmark$ &$\checkmark$ &68.84 &74.56 &60.68 &72.83 &78.76 &77.85 &81.44 &83.12 \\ \hline
\end{tabular}
\label{tab:imagnet}
\end{table*}

The Group Quantization Unit supports efficient processing of post-activation values by implementing a three-stage, group-wise quantization architecture. Based on offline calibration, the unit first selects a base group in the Quantization Scale Allocation stage, deriving its reference scale from statistical features, while estimating scales for other groups via shift-based approximations. In the Intra-group Quantization stage, channels within each group—having similar value distributions—share integer-only quantization parameters. Dynamic bit-width adjustment is applied to balance quantization precision and hardware cost. Finally, in the Cross-group Alignment stage, to support matrix accumulation across groups with different quantization scales, inverse shifting is applied to non-base groups, aligning all values using bit-shifting alone.

As illustrated in Figure~\ref{fig:group}, this hardware-friendly design avoids floating-point operations, relying instead on simple shift and add primitives. The architecture supports parallel group processing and synergizes with other modules in the \Alg framework to ensure low energy consumption and high throughput without sacrificing quantization accuracy.

\begin{table*}[htbp]
    \setlength{\abovecaptionskip}{3pt}
    \setlength{\belowcaptionskip}{3pt}
\centering
\footnotesize
\vspace{-0.5cm}
\caption{PTQ performance on GLUE benchmark.}
\vspace{-0.2cm}
\resizebox{\textwidth}{!}{%
\begin{tabular}{@{}cccccccccccccc@{}}
\toprule
\multirow{2}{*}{\textbf{Method}} & \textbf{Prec.} & \textbf{CoLA} & \textbf{MNLI} & \textbf{MRPC} & \textbf{QNLI} & \textbf{QQP} & \textbf{RTE} & \textbf{SST-2} & \textbf{STS-B} & \multirow{2}{*}{\textbf{Avg.}} \\ 
                & (W/A)       & (Matt.)       & (acc/mm)      & (f1/acc)      & (acc)         & (f1/acc)    & (acc)       & (acc)         & (Pear./Spear.) &       \\ 
\midrule
{\textbf{BERT}} &32-32 &59.60 &84.94/84.76 & 91.35/87.75   & 91.84         & 87.82/90.91 & 72.56       & 93.35         & 89.70/89.22    & 83.83 \\ 
\midrule
Percentile\cite{mckinstry2018discovering}        & 6-6         & 37.32         & 72.40/71.69   & 85.09/79.90   & 79.37         & 72.58/80.19 & 61.73       & 87.27         & 86.38/87.29    & 72.93 \\ 
EasyQuant\cite{wu2020easyquant} & 6-6         & 38.16         & 75.82/75.66   & 82.51/77.45   & 84.94         & 75.31/81.81 & 65.34       & 87.27         & 85.50/86.33    & 74.49 \\ 
Outlier Sup.\cite{wei2022outlier}  & 6-6         & 54.40         & 82.02/81.69   & 87.45/83.33   & 89.82         & 84.69/88.94 & 70.76      & 91.86         & 88.65/88.55    & 81.19 \\ 
 \rowcolor{green!15} 
\textbf{QUARK(ours)}            & 6-6         & \textbf{61.41}         &\textbf{82.78/81.61}    &\textbf{88.19/83.56}    & \textbf{90.20}         &\textbf{86.88/89.83}  & \textbf{70.15}       & \textbf{91.09}         & \textbf{88.38/87.50}    &\textbf{82.25}  \\ \hline

Percentile\cite{mckinstry2018discovering}     & 4-4         & -4.15         & 32.73/32.95   & 81.22/68.38   & 50.36         & 23.16/60.71  & 47.29       & 50.92         & -9.18/-8.63    & 35.64 \\ 
Outlier Sup.\cite{wei2022outlier}  & 4-4         & -1.63         & 27.09/26.55   & 68.00/60.78   & 59.64         & 23.34/65.87  & 52.71       & 70.41         & 0.48/-0.29    & 39.50 \\ 
 \rowcolor{green!15} 
\textbf{QUARK(ours)}             & 4-4         & \textbf{43.76}         & \textbf{39.15/40.64}   & \textbf{80.43/67.12}   &  \textbf{67.08}        &\textbf{35.99/72.76}  & \textbf{51.77}       & \textbf{84.39}         & \textbf{63.67/64.82}    & \textbf{59.91} \\ 
\midrule
{\textbf{RoBERTa}}  &32-32 &62.50 &87.75/87.23 & 93.10/90.44   & 92.68         & 88.78/91.60 & 80.51       & 95.18         & 91.04/90.72    & 86.40 \\ 
\midrule

Percentile\cite{mckinstry2018discovering}            & 6-6         & 20.73         & 72.23/73.68   & 84.43/78.43  & 77.16         & 82.21/87.44 & 62.82       & 88.19          & 79.41/79.64    & 70.98 \\ 
EasyQuant\cite{wu2020easyquant}      & 6-6         & 9.28         & 74.96/75.87   & 84.31/76.47  & 74.04      & 85.52/89.12 & 62.45       & 89.56         & 80.89/82.38    & 70.01 \\ 
Outlier Sup.\cite{wei2022outlier}   & 6-6         & 46.35         & 83.38/83.32   & 87.50/83.33  & 86.82     & 86.86/90.01 & 67.51       & 92.20        & 86.83/86.93    & 79.62 \\ 
 \rowcolor{green!15} 
\textbf{QUARK(ours)}            & 6-6         &  \textbf{49.49}        & \textbf{83.33/82.68}   & \textbf{87.65/83.42}   & \textbf{86.98}         &\textbf{86.34/89.03}  & \textbf{78.75}       &  \textbf{92.72}        & \textbf{90.02/89.92}    & \textbf{81.77} \\ \hline

Percentile\cite{mckinstry2018discovering}    & 4-4         & 0.01         & 31.82/31.72   & 77.22/63.38   & 50.54         & 38.15/52.92  & 47.29       & 50.92         & -1.52/-0.99    & 36.89 \\ 
Outlier Sup.\cite{wei2022outlier}  & 4-4         & -3.24         & 32.75/32.37   & 76.76/63.84   & 49.5         & 15.75/60.36  & 52.35       & 51.15         & -3.15/-2.52    & 35.98 \\ 
 \rowcolor{green!15} 
\textbf{QUARK(ours)}            & 4-4         &  \textbf{10.49}        & \textbf{36.26/37.33}   & \textbf{75.44/66.72}   &  \textbf{52.45}        & \textbf{40.75/58.39} & \textbf{73.95}       & \textbf{90.28}         & \textbf{88.71/87.03}    & \textbf{59.02} \\ 
\bottomrule
\end{tabular}%
}
\label{tab:quantization_results}
\vspace{-0.5cm}
\end{table*}

\section{Experiments}
\subsection{Experiment Setting}
\textbf{Software Setup}
To validate our algorithm, we conducted experiments on both CV and NLP tasks. For CV tasks, we evaluated image classification on the ImageNet\cite{deng2009imagenet} dataset with different model variants, including ViT\cite{dosovitskiy2021imageworth16x16words}, DeiT\cite{touvron2021training}, and Swin Transformer\cite{liu2021swin}. For NLP tasks, we performed experiments on the GLUE\cite{wang2018glue} benchmark using BERT-Base\cite{devlin2019bert} and RoBERTa-Base\cite{liu2019roberta} models. To ensure fair comparisons with prior works, we randomly selected 32 samples from the ImageNet dataset for image classification tasks and 256 samples in NLP tasks to calibrate quantization parameters.

\textbf{Hardware Setup}
To evaluate the impact of the proposed method on hardware overhead, we synthesized and implemented the approach on a ZCU102 board using Vivado. At a clock frequency of 300 MHz, the hardware resource utilization was obtained from the place-and-route results. The results were then compared with existing FPGA implementations.

\subsection{Accuracy Results}
\textbf{a) Image Classification on ImageNet:}
As shown in Table \ref{tab:imagnet}, our method outperforms I-BERT under W8A8 quantization and nearly matches I-ViT ($<1\%$ accuracy loss), despite I-ViT requiring costly retraining. For W6A6, our approach surpasses FQ-ViT and PTQ4ViT and performs comparably to RepQ-ViT. Unlike PTQ methods, our integer-only framework eliminates hardware challenges. For W4A4, we set a new state-of-the-art, achieving a 6.08\% accuracy gain on ViT-B and $>3\%$ average improvement across various networks. 

\textbf{b) Language Understanding on GLUE:}
Table \ref{tab:quantization_results} summarizes our results on the GLUE benchmark. At 6-bit quantization, \Alg achieves significant improvements, e.g., 7.01\% gain for BERT on CoLA. At 4-bit, \Alg outperforms prior methods, improving BERT's average GLUE score by 20.41\% and RoBERTa's by 23.04\%. These results demonstrate the robustness of \Alg for language tasks.

\begin{figure}[t]

    \setlength{\abovecaptionskip}{3pt}
    \setlength{\belowcaptionskip}{3pt}
    \centering
    \includegraphics[width=\columnwidth]{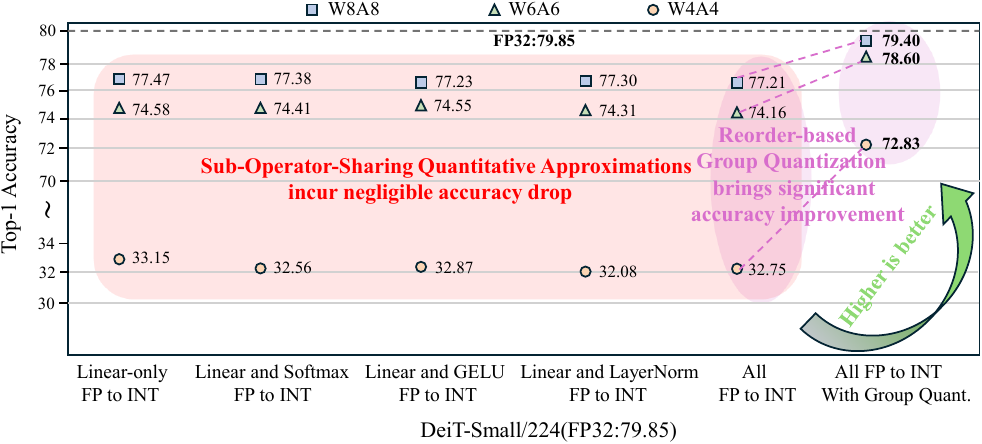}
    \vspace{-0.6cm}
    \caption{The impact of individual nonlinear operators (Softmax, GELU, and LayerNorm) and the Reorder-based Group Quantization on model accuracy across different bit configurations (W8A8, W6A6, W4A4).}
    \label{fig:ablation}
    \vspace{-0.3cm}
\end{figure}

\begin{table*}[t!]
\setlength{\abovecaptionskip}{3pt}
\setlength{\belowcaptionskip}{3pt}
\centering
\caption{Performance Comparison With Previous Works.}
\label{tab:fpga_compare}
\begin{tabular}{ccccccccc}
\hline
\multirow{2}{*}{\textbf{Method}} &\textbf{FPGA} & \multirow{2}{*}{\textbf{Model}} & \textbf{Prec.} & \textbf{DSP} & \textbf{LUT} & \multirow{2}{*}{\textbf{Freq.}}  & \multirow{2}{*}{\textbf{Latency}} & \multirow{2}{*}{\textbf{Throughput}}  \\ 
 &\textbf{Platform} & &\textbf{(W/A)} &\textbf{Utilization} &\textbf{Utilization} &&& \\
 \hline
 ViA\cite{wang2022via} &Alveo U50 &Swin-T &F16 &2420/5952(40.6\%) &258/872(29.6\%) &300MHz &- &309.6GOPs \\
 Auto-ViT-Acc~\cite{sun2022fpga} &ZCU102 &DeiT-S &W8/4 A8 &1552/2520(61.6\%) &185/274(67.5\%) &150MHz &- &907.8GOPs \\
TCAS-I'23~\cite{10288182} &ZCU102 &ViT-S &W8A8 &1268/2520(50.3\%) &144/274(52.7\%) &300MHz &11.15ms &762.7GOPs \\
 \rowcolor{green!15} 
 QUARK(ours) &ZCU102 &ViT-S &W8A8 &1181/2520(46.9\%) &130/274(47.4\%) &300MHz &10.68ms &787.5GOPs \\
 HotaQ\cite{10737419} &ZCU102 &ViT-S &W4A4 &1933/2520(76.7\%) &157/274(57.3\%) &200MHz &7.59ms &733.7GOPs \\
  \rowcolor{green!15} 
 QUARK(ours) &ZCU102 &ViT-S &W4A4 &1056/2520(41.9\%) &147/274(53.6\%) &300MHz &6.59ms &864.7GOPs \\
\hline
\end{tabular}
\vspace{-0.5cm}
\end{table*}

\textbf{c) Ablation study:} We evaluate the effectiveness of nonlinear operator approximation and reordering-based group quantization across various bit-widths, using linear-layer-only quantization as the baseline. As shown in Figure \ref{fig:ablation}, individual nonlinear operator approximation causes minimal accuracy loss, with an average drop below 0.2\% for W8A8 and W6A6 quantization. When applying three approximations simultaneously, the accuracy loss remains at 0.26\% for W8A8 and 0.42\% for W6A6, and only 0.4\% for W4A4 quantization on the DeiT-Small model. Notably, our reordering-based group quantization not only compensates for approximation errors but also significantly enhances quantization precision. Compared to full-precision models, it achieves only 0.45\% accuracy loss under W8A8 quantization and delivers over 40\% accuracy improvement in ultra-low-bit quantization (W4A4).

\subsection{Hardware Evaluation}
\textbf{a) Overall Performance:}
Table~\ref{tab:fpga_compare} shows significant advancements over SOTA approaches. Our design achieves a peak throughput of 787.5 GOP/s with minimal hardware overhead, outperforming existing solutions across multiple dimensions.
Compared to HotaQ\cite{10737419}, which employs hardware-oriented token adaptive
quantization for embedded FPGAs, our approach delivers superior network accuracy despite maintaining higher hardware efficiency. While ViA\cite{wang2022via} optimizes data locality through ViT structure analysis and partitioning strategies, its FP16-based implementation achieves only 309.6 GOP/s, significantly lower than our design. Auto-ViT-Acc\cite{sun2022fpga} surpasses our throughput through mixed-precision quantization including PoT formats, but this approach compromises accuracy and introduces substantial training overhead. 

\textbf{b) Nonlinear Operators Resource Utilization:}
As shown in Table \ref{tab:hardware}, our design achieves significant hardware resource savings compared to conventional solutions, reducing LUT utilization to 20\%-80\% and FF consumption by approximately 70\%. Traditional implementations rely on memory-intensive LUT methods or extensive division operations. In contrast, our design leverages the TDM features of nonlinear operators across Vision Transformer stages to enable dynamic resource sharing. Specifically, we introduce a time-multiplexed shared circuit that dynamically reuses computational resources across processing stages, eliminating the hardware redundancy incurred by dedicated circuits for individual nonlinear operators.

\begin{table}[h]
\vspace{-0.3cm}
\centering

\caption{Comparison of hardware overhead with SOTA FPGA implementations. \textbf{*} indicates the resource consumption of non-shared components, excluding shared resources in nonlinear operators.}
\vspace{-0.3cm}
\setlength{\tabcolsep}{3pt}
\resizebox{1.0\linewidth}{!}{
\begin{tabular}{ccccccc}
\hline
\multirow{2}{*}{\textbf{Method}}  & \multirow{2}{*}{\textbf{Platform}}     & \multirow{2}{*}{
\begin{tabular}[l]{@{}c@{}}\textbf{Hardware}\\\textbf{Resource} \end{tabular}
} 
& \multicolumn{4}{c}{\textbf{Resource Utilization}} \\ \cline{4-7} 
& & &  \textbf{Softmax} & \textbf{GELU} & \textbf{LayerNorm} & \textbf{Total} \\ 
\hline
\multirow{3}{*}{Swat\cite{10473931}} &\multirow{3}{*}{Alevo U50}    & \cellcolor{myblue!30}LUT                           & \cellcolor{myblue!30}135,093           & \cellcolor{myblue!30}11,219         & \cellcolor{myblue!30} 10,731             &\cellcolor{myblue!30} 157,043         \\ &
                       & \cellcolor{mygreen!30}FF                            & \cellcolor{mygreen!30}-                &\cellcolor{mygreen!30} -             & \cellcolor{mygreen!30}-                 &\cellcolor{mygreen!30} -              \\ &
                       & \cellcolor{myorange!30}DSP                           & \cellcolor{myorange!30}98               & \cellcolor{myorange!30}196           &\cellcolor{myorange!30} 0                 & \cellcolor{myorange!30}294            \\ \hline
\multirow{3}{*}{ViA\cite{wang2022via}}  &\multirow{3}{*}{Alevo U50} &\cellcolor{myblue!30} LUT                           & \cellcolor{myblue!30}32,783            & \cellcolor{myblue!30}3,325          & \cellcolor{myblue!30}2,817              & \cellcolor{myblue!30}38,925          \\ &  
                       & \cellcolor{mygreen!30}FF                            & \cellcolor{mygreen!30}32,383            & \cellcolor{mygreen!30}3,161          & \cellcolor{mygreen!30}2,145              & \cellcolor{mygreen!30}37,689          \\  &
                       & \cellcolor{myorange!30}DSP                           & \cellcolor{myorange!30}131              & \cellcolor{myorange!30}20            &\cellcolor{myorange!30} 7                 & \cellcolor{myorange!30}158            \\ \hline
\multirow{3}{*}{SOCC'20\cite{9524802}} &\multirow{3}{*}{
\begin{tabular}[l]{@{}c@{}}Virtex UltraScale\\+ VXU13P \end{tabular}
} & \cellcolor{myblue!30}LUT                  & \cellcolor{myblue!30}21,190            &\cellcolor{myblue!30} -             & \cellcolor{myblue!30}10,551             & \cellcolor{myblue!30}-          \\ &
                       & \cellcolor{mygreen!30}FF                            &\cellcolor{mygreen!30} 32,623            &\cellcolor{mygreen!30} -             & \cellcolor{mygreen!30}5,325              & \cellcolor{mygreen!30}-              \\  &
                       & \cellcolor{myorange!30}DSP                           &\cellcolor{myorange!30} 0                & \cellcolor{myorange!30}-             & \cellcolor{myorange!30}129               &\cellcolor{myorange!30} -              \\ \hline
\multirow{3}{*}{TCAS-I'23\cite{10288182}} &\multirow{3}{*}{ZCU102} & \cellcolor{myblue!30}LUT                   &\cellcolor{myblue!30} 22,865            &\cellcolor{myblue!30} 10,163         & \cellcolor{myblue!30}10,558             & \cellcolor{myblue!30}43,586          \\ &
                       & \cellcolor{mygreen!30}FF                            & \cellcolor{mygreen!30}21,770            & \cellcolor{mygreen!30}5,992          & \cellcolor{mygreen!30}4,038              & \cellcolor{mygreen!30}31,800          \\  &
                       & \cellcolor{myorange!30}DSP                           & \cellcolor{myorange!30}128              &\cellcolor{myorange!30} 32            &\cellcolor{myorange!30} 7                 & \cellcolor{myorange!30}167            \\ \hline
\multirow{3}{*}{QUARK(ours)} &\multirow{3}{*}{ZCU102} & \cellcolor{myblue!30}LUT                           &\cellcolor{myblue!30} 21,804            & \cellcolor{myblue!30}2,772*          & \cellcolor{myblue!30}4,959*              &\cellcolor{myblue!30} \textbf{29535}          \\ &
                       & \cellcolor{mygreen!30}FF                            & \cellcolor{mygreen!30}8,184             & \cellcolor{mygreen!30}1,815*         & \cellcolor{mygreen!30}1,015*                 & \cellcolor{mygreen!30}\textbf{11,014}          \\ &
                       & \cellcolor{myorange!30}DSP                           & \cellcolor{myorange!30}144              & \cellcolor{myorange!30}0*            & \cellcolor{myorange!30}3*                & \cellcolor{myorange!30}\textbf{147}            \\ \hline
\end{tabular}
}
\label{tab:hardware}
\end{table}

\begin{figure}[t!]
\setlength{\abovecaptionskip}{3pt}
\setlength{\belowcaptionskip}{3pt}
    \centering
    \includegraphics[width=\columnwidth]{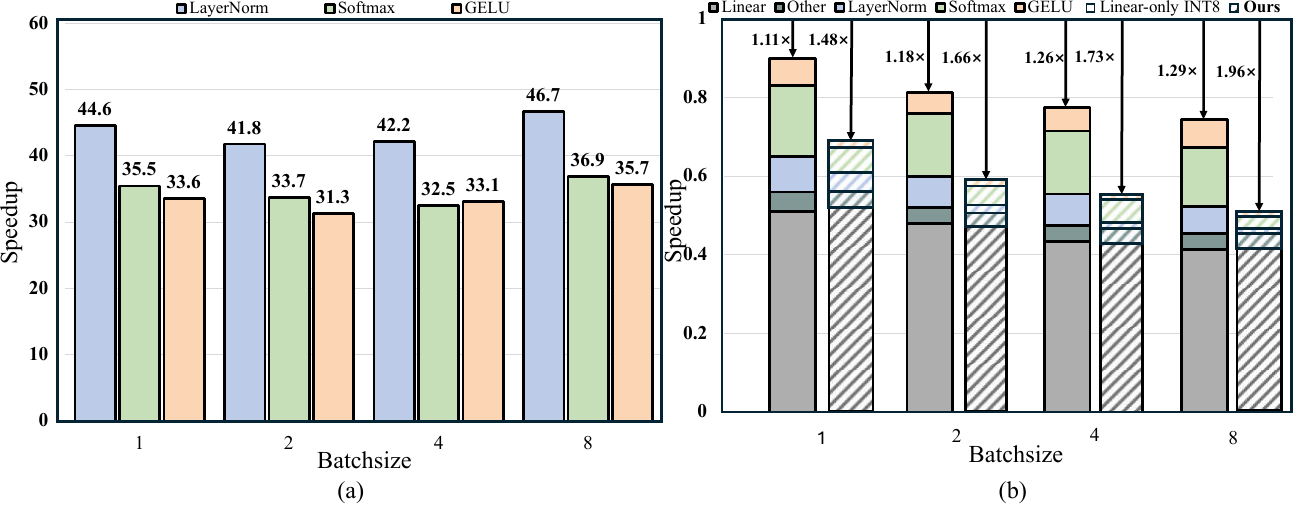}
    \vspace{-0.5cm}
    \caption{(a) Speedup over GPU on all nonlinear operators. (b) End-to-end speedup and latency breakdown, the latency is normalized with respect to FP32 implementation.}
    \label{fig:operators}
\end{figure}

\textbf{c) Latency Comparison:}
Figure \ref{fig:operators} (a) shows QUARK's performance on the ZCU102 FPGA platform compared to RTX A5000 GPU, evaluating all nonlinear operators in DeiT-Tiny (batch sizes 1-8, token length 197). \Alg achieves substantial speedups of 41.8×-46.7× for LayerNorm, 32.5×-36.9× for Softmax, and 31.3×-35.7× for GELU, enabled by hardware-friendly approximations and optimized pipelined datapath designs. Figure \ref{fig:operators} (b) reveals that while INT8 models with linear layer-only quantization show limited GPU speedups (1.11×-1.29×) due to dominant nonlinear operations, QUARK's nonlinear operator optimization enables remarkable 1.48×-1.96× end-to-end speedup.

\section{Conclusion}
This work presents QUARK, a software/hardware co-design solution for accelerating quantized Transformer-based models on FPGA platforms. By integrating lightweight integer-only arithmetic modules and a reorder-based group quantization. Evaluations show that \Alg delivers state-of-the-art efficiency gains for vision and language Transformers under various integer-only inference settings.

\bibliography{ref}
\bibliographystyle{IEEEtran}

\end{document}